\documentclass[10pt,twocolumn,letterpaper]{article}

\usepackage{multirow}
\usepackage[pagenumbers]{cvpr} % To force page numbers, e.g. for an arXiv version

\definecolor{cvprblue}{rgb}{0.21,0.49,0.74}
\usepackage[pagebackref,breaklinks,colorlinks,allcolors=cvprblue]{hyperref}

\def\paperID{*****} % *** Enter the Paper ID here
\def\confName{CVPR}
\def\confYear{2027}

\title{CrossFit-GS: Out-of-Fold Pose Estimation for COLMAP-Free 3D Gaussian Splatting}

\author{
Zijian Wu \quad Mingfeng Jiang \quad Zidian Lin \quad Ying Song \quad Ziqian Lu\\
Zhejiang Sci-Tech University\\
{\tt\small} % 选填：可以放邮箱
\and
Jinliang Wang\\
Tsinghua University\\
{} % 选填：可以放邮箱
}

\begin{document}

\twocolumn[{%
\renewcommand\twocolumn[1][]{#1}%
\maketitle
\begin{center}
    \captionsetup{type=figure}
    \includegraphics[width=\textwidth]{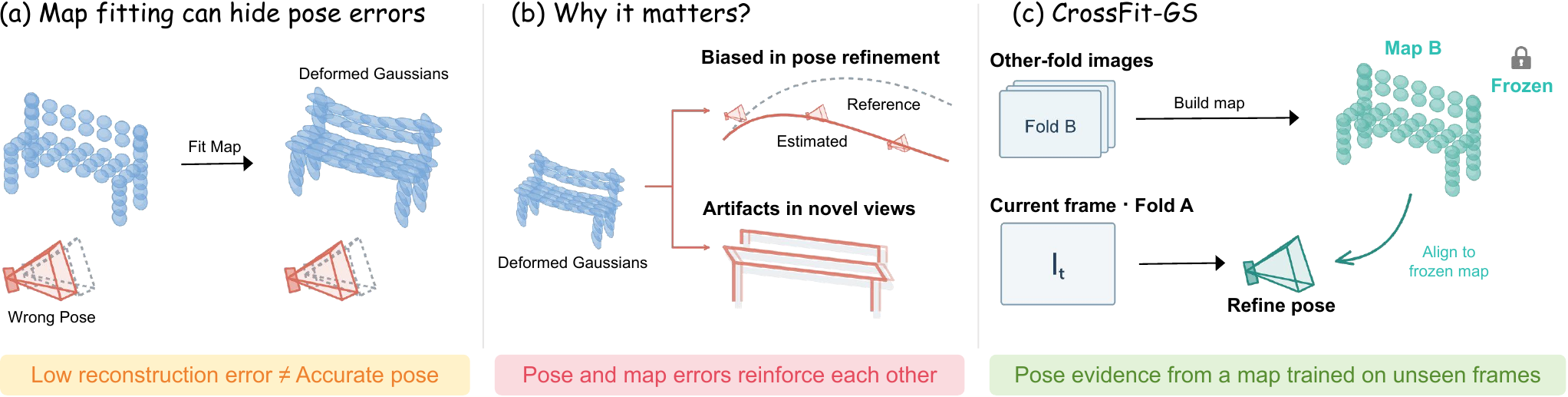}
    \captionof{figure}{\textbf{Pose--map compensation motivates cross-fitted camera registration.} (a) Gaussian deformation can reduce training-view error while an incorrect camera pose persists. (b) The resulting geometry can bias subsequent registration and introduce novel-view artifacts. (c) CrossFit-GS refines each frame's pose against a frozen Gaussian map trained on the opposite frame partition, alternating the roles of the two partitions.}
    \label{fig:teaser}
\end{center}%
%\vspace{0.5cm} % 稍微控制一下 Teaser 图与下面 Abstract 的间距
}]

\maketitle

\begin{abstract}
Pose-free 3D Gaussian Splatting (3DGS) aims to bypass cumbersome Structure-from-Motion (SfM) by directly reconstructing scenes from unposed image sequences. In this work, we present CrossFit-GS, a deferred-commitment progressive framework that mitigates cumulative trajectory drift and breaks the self-reinforcing loop of distorted-map error propagation. Specifically, we partition the input sequence into interleaved subsets and maintain two complementary submaps. Each candidate pose is cross-refined against a frozen opposite submap that has never been optimized using the target frame, thereby providing an independent geometric reference for pose estimation. Checkerboard validation and spatial coverage safeguards further reject unreliable updates, and only verified poses are committed to their corresponding owner submap. Once the trajectory has been refined, the intermediate submaps are discarded and the final global representation is trained from scratch using the fixed refined poses. Extensive evaluations across multiple benchmarks demonstrate that CrossFit-GS substantially reduces cumulative trajectory drift and consistently improves both camera pose estimation and novel-view synthesis over existing methods.
\end{abstract}    
\section{Introduction}
\label{sec:intro}

3D Gaussian Splatting (3DGS) \cite{kerbl20233d,10870258} has demonstrated remarkable success across a wide range of applications \cite{qian20243dgs, liu2024citygaussian,c3dgs,11093421}. However, standard 3DGS heavily relies on pre-computed camera extrinsics from classical Structure-from-Motion (SfM) pipelines like COLMAP \cite{colmap}. However, SfM preprocessing is computationally expensive, sensitive to rapid camera motion, and prone to failure in texture-less or low-overlap regions, hindering the practical deployment of 3DGS in real-time or streaming scenarios \cite{huang20253r, guo2025robust, ICLR2025_bf11e0bb}.

Recent efforts explore pose-free reconstruction paradigms that estimate camera trajectories and scene geometry directly from unposed image sequences \cite{NEURIPS2025_7cdc4122, 10204053, Lin_2025_ICCV, dong2025betterrobustnessposefree3d}. Initial implicit methods \cite{SCNeRF2021, nerfmm, 10.1609/aaai.v38i6.28464}, such as BARF \cite{barf} and Nope-NeRF \cite{nope-nerf}, pioneered this direction by fitting coordinate-based Neural Radiance Fields (NeRFs) \cite{mildenhall2021nerf} alongside camera frame parameters. Leveraging the explicit representation and fast rasterization of Gaussians, frameworks like CF-3DGS \cite{cf-3dgs} adapted this concept into a sequential regime by estimating relative poses with local Gaussian models while dynamically expanding a global scene representation. Follow-up studies improve complementary components of this pipeline by enhancing Gaussian coverage through hierarchical merging \cite{ji2025sfm}, refining local pose estimation via feature and frequency regularization \cite{wei2025pcr}, and introducing global trajectory constraints using long-range feature tracks \cite{shi2026trackgs}.

Despite these empirical advances, existing progressive paradigms fundamentally struggle to reconcile local tracking tractability with global geometric consistency. Standard sequential frameworks estimate pairwise relative transformations from local Gaussian representations and compose them into a global trajectory \cite{cf-3dgs,wei2025pcr,ji2025sfm}. Although this localized formulation makes pose tracking tractable, residual pairwise errors can accumulate along the sequence and are absorbed by the global scene representation \cite{Huang_2025_ICCV,Mao_2025_ICCV}. Spatially flexible Gaussians can deform to accommodate these registration offsets, yielding deceptively low training-view reconstruction errors despite inaccurate camera trajectories \cite{huang20253r,Kong_2026_CVPR}. As additional frames are incorporated, the evolving representation is repeatedly optimized to explain new observations under accumulated pose errors, causing geometric distortion to propagate throughout the reconstruction. While scene-level pose refinement offers a theoretical remedy for such cumulative drift, optimizing a camera pose against a scene representation trained on that very same observation creates a self-explanation shortcut \cite{schmidt2024lookgaussposenovel, Peng_2025_ICCV, Truong_2023_CVPR}. Unresolved, this underlying dilemma often leads existing methodologies to focus heavily on refining initial pose estimates, rather than addressing systemic trajectory drift at its root. We highlight this self-explanation loop as a critical bottleneck in progressive camera-free reconstruction.

To overcome this fundamental limitation, we propose CrossFit-GS, a progressive pose-free reconstruction framework built on a deferred-commitment protocol. Centered on dual-map information-flow isolation, CrossFit-GS strictly decouples camera pose refinement from scene map updates for unposed image streams. Specifically, the input sequence is partitioned into two interleaved temporal subsets ($A/B$ folds), each maintaining an independent Gaussian submap. For an incoming frame, a candidate pose proposal is generated from the preceding motion estimate and cross-refined exclusively against the frozen opposite submap. To improve optimization stability, this cross-map pose refinement is strictly regulated by checkerboard validation masks and spatial coverage gates. Following our deferred-commitment principle, candidate pose updates are committed to the global trajectory and integrated into owner submap updates only after passing these validation checks. Once all camera extrinsics across the sequence are verified and frozen, the intermediate submaps are discarded, and a single global Gaussian representation is initialized and trained from scratch under the fixed trajectory. Through these coordinated strategies, CrossFit-GS effectively breaks the self-reinforcing pose--map error propagation loop, substantially mitigating cumulative trajectory drift while yielding highly accurate camera estimates and superior novel-view rendering quality.

Our contributions are summarized as follows:

\begin{itemize}
    \item We diagnose a fundamental bottleneck in COLMAP-free 3DGS, highlighting how accumulated pairwise errors distort the evolving scene representation and how reusing this pose-contaminated map for scene-level refinement creates a self-explanation shortcut that closes a self-reinforcing pose--map fitting loop.
    
    \item We design a deferred-commitment progressive framework, CrossFit-GS, that prevents error propagation and suppresses trajectory drift by explicitly decoupling pose refinement from scene reconstruction through dual-map information isolation and cross-validation safeguards.
    
    \item Extensive experiments validate that CrossFit-GS significantly suppresses trajectory drift, achieving state-of-the-art camera tracking and view synthesis.
\end{itemize}
\begin{figure*}[t]
    \centering
    \includegraphics[width=\textwidth]{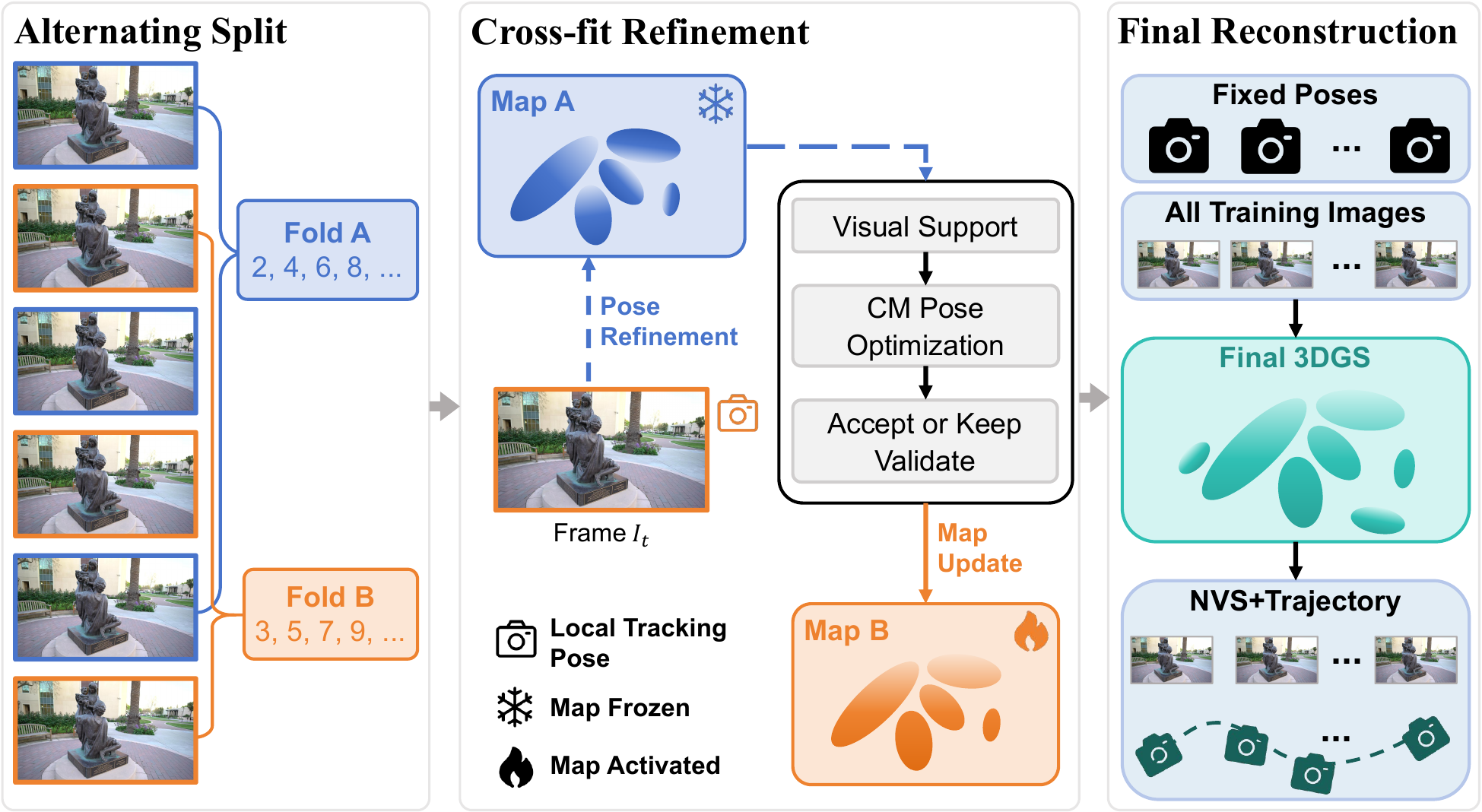}
    \caption{
        \textbf{Overview of CrossFit-GS.}
        The input sequence is partitioned into two interleaved folds,
        each associated with a separate Gaussian map.
        For a frame in Fold B, its pose is refined against the frozen
        Map A, with visual-support checks and an accept-or-keep
        validation rule governing the update.
        The frame subsequently contributes only to Map B.
        The roles of the two folds alternate, ensuring that a frame
        does not train the map used to assess its pose.
        Finally, a unified Gaussian representation is trained from
        scratch using all training images and the fixed estimated poses.
    }
    \label{fig:pipeline}
\end{figure*}

\section{Related Work}

\subsection{Generalizable Feed-Forward Methods}
One line of research addresses unposed sparse-view reconstruction using
generalizable models \cite{fang2026dens3r, wang2025vggt} learned from large-scale multi-view data \cite{yu2021pixelnerf,Charatan_2024_CVPR,feng2025flashgs, venkatraman2026vggt, szymanowicz2024splatter, Huang_2025_ICCV}.
FreeSplatter \cite{Xu_2025_ICCV} predicts pixel-aligned Gaussian primitives
in a shared reference frame and recovers camera parameters from the
predicted geometry. SPFSplat \cite{Huang_2025_ICCV} jointly predicts camera
poses and 3D Gaussians from unposed sparse views in a single feed-forward
pass without pose supervision. Related approaches combine learned
multi-view geometric priors with per-scene optimization. InstantSplat
\cite{fan2024instantsplat}, for example, initializes camera poses and dense
geometry using a pretrained multi-view model and subsequently refines the
scene through Gaussian bundle adjustment.

These methods are primarily designed for rapid reconstruction from a small
set of sparse, potentially unordered views. Their learned priors provide
strong geometric initialization but introduce dependence on the training
distribution and are not specifically formulated for incrementally
processing long, temporally ordered sequences. Our work instead considers
per-scene progressive reconstruction, where camera poses must remain
globally consistent as observations are incorporated throughout an extended
sequence.

\subsection{Sequential Pose-Free Reconstruction}
Sequential COLMAP-free methods exploit temporal continuity to estimate camera motion and scene geometry through per-scene optimization \cite{Lin_2025_ICCV, Gao_2026_CVPR}. Early implicit methods achieve pose-free reconstruction by jointly optimizing camera parameters alongside radiance fields \cite{yang2023nerfvs, 10.1145/3664647.3681507, wang2023f2nerf}, employing frequency annealing in BARF \cite{barf} and monocular depth consistency in Nope-NeRF \cite{nope-nerf} to prevent optimization collapse. Transitioning to explicit representations \cite{sun2024correspondence, chen2024zerogs}, CF-3DGS \cite{cf-3dgs} estimates relative transformations between adjacent frames using local Gaussian models, composes them into a camera trajectory, and progressively grows a global Gaussian representation. Subsequent work improves different components of this pipeline. PCR-GS \cite{wei2025pcr} introduces feature-reprojection and frequency-domain \cite{oquab2023dinov2} regularization to improve local pose estimation. HT-3DGS \cite{ji2025sfm} uses video frame interpolation to reduce the motion between adjacent observations and hierarchically trains and merges segment-specific Gaussian models to improve scene coverage. TrackGS \cite{shi2026trackgs} incorporates long-range feature tracks, track-constrained Gaussians, and multi-view geometric losses to jointly optimize camera parameters and the Gaussian representation under global trajectory constraints.

Despite these advances, sequential methods remain vulnerable to cumulative drift from locally composed pose estimates, whose errors can be absorbed by
the flexible Gaussian representation. When long-range refinement is introduced, poses are typically optimized against a map trained on the same observations, leaving a self-explanation pathway between pose estimation and
map fitting.

\section{Method}
\label{sec:method}

\subsection{Problem Formulation and System Overview}
\label{sec:overview}
Given an unposed sequence of RGB images $\{I_i\}_{i=0}^{N-1}$ with known intrinsic matrix $K$, our goal is to recover the camera trajectory $\{T_i \in \mathrm{SE}(3)\}_{i=0}^{N-1}$ and subsequently reconstruct a global 3D Gaussian scene representation $\mathcal{G} = \{(\boldsymbol{\mu}_j, \boldsymbol{\Sigma}_j, \alpha_j, \mathbf{f}_j)\}_{j=1}^{M}$~\cite{kerbl20233d}. Differentiable rasterization renders a synthesized RGB image $\mathcal{R}(\mathcal{G}, T_i; K)$ along with an accumulated opacity map $A(\mathcal{G}, T_i; K)$.

To reconcile local tracking with global geometric consistency and suppress cumulative drift, CrossFit-GS introduces an observation-isolated pipeline under a deferred-commitment protocol (Fig.~\ref{fig:pipeline}). Rather than continuously updating a shared global map during pose estimation, our framework decouples trajectory solving from global scene synthesis through three functional components: \textbf{(i) Alternating Split and Local Tracking}, which streamwise partitions incoming frames into interleaved temporal folds and derives initial pose proposals; \textbf{(ii) Cross-fit Refinement and Validation}, which optimizes target camera poses against the frozen opposite submap while updating the activated owner submap only upon passing validation controls; and \textbf{(iii) Final Reconstruction under Deferred Commitment}, which discards intermediate submaps once all camera extrinsics are verified and frozen, utilizing the fixed poses to train a unified global 3DGS representation from scratch.

\subsection{Dual-Map Partitioning and Local Pose Tracking}
\label{sec:split_tracking}

To enforce observation isolation during progressive tracking, the input frame stream $\{I_i\}_{i=0}^{N-1}$ is temporally partitioned according to index parity, with owner assignment $a(i)=i\bmod 2$ and opposite assignment $b(i)=1-a(i)$. We initialize the progressive reconstruction from the first frame, which defines the world coordinate system with $\widehat{T}_0=\mathbf{I}$ and provides the monocular-depth initialization of the Gaussian representation. As subsequent frames arrive, they are assigned to their owner folds, yielding two source image sets $\mathcal{S}_0$ and $\mathcal{S}_1$ and their corresponding provisional Gaussian submaps $\mathcal{G}^{(0)}$ and $\mathcal{G}^{(1)}$. The two submaps are constructed and updated alternately throughout the sequence.

For each target frame $I_i$, the opposite submap used for pose refinement excludes that frame from its source image set:
\begin{equation}
    I_i \notin \mathcal{S}_{b(i)}.
    \label{eq:exclusion}
\end{equation}

This exclusion ensures that the representation used to refine $T_i$ has received no direct training feedback from $I_i$, thereby preventing direct target-to-map self-fitting.

When a new frame $I_i$ arrives ($i \ge 1$), a local tracker first provides an initial pose proposal $\widetilde{T}_i \in \mathrm{SE}(3)$ before cross-map refinement. Specifically, a reference Gaussian model is constructed from the preceding frame $I_{i-1}$ using monocular depth estimation. Differentiable rendering against this local model estimates a relative motion $\widehat{\Delta}_i$, initialized via short-term kinematic continuity as $\Delta_i^{(0)} = \widehat{T}_{i-1} \widehat{T}_{i-2}^{-1}$ for $i \ge 2$ (and identity $\mathbf{I}$ for $i = 1$). The local pose proposal is formulated as:
\begin{equation}
    \widetilde{T}_i = \widehat{\Delta}_i \widehat{T}_{i-1}.
    \label{eq:localtracking}
\end{equation}
Because $\widetilde{T}_i$ relies solely on a single frame's unrefined local geometry, it serves strictly as an initial proposal. To verify and refine $\widetilde{T}_i$, target frame $I_i$ is subsequently registered against the frozen opposite submap $\mathcal{G}^{(b(i))}$, where all Gaussian parameters are detached from gradient computation to optimize only $T_i$.

\begin{table*}[t]
\centering
\small
\caption{Quantitative novel view synthesis results on Tanks and Temples. Evaluated using PSNR~$\uparrow$, SSIM~$\uparrow$, and LPIPS~$\downarrow$. Best results are bolded.}
\label{tab:tanks_nvs}
\setlength{\tabcolsep}{3.5pt}
\begin{tabular}{l|ccc|ccc|ccc|ccc|ccc}
\toprule
\multirow{2}{*}{Scene} & \multicolumn{3}{c|}{CF-3DGS~\cite{cf-3dgs}} & \multicolumn{3}{c|}{HT-3DGS~\cite{ji2025sfm}} & \multicolumn{3}{c|}{TrackGS~\cite{shi2026trackgs}} & \multicolumn{3}{c|}{PCR-GS~\cite{wei2025pcr}} & \multicolumn{3}{c}{\textbf{Ours}} \\
& PSNR & SSIM & LPIPS & PSNR & SSIM & LPIPS & PSNR & SSIM & LPIPS & PSNR & SSIM & LPIPS & PSNR & SSIM & LPIPS \\
\midrule
Ballroom & 24.917 & 0.828 & 0.123 & 25.620 & 0.850 & 0.110 & 25.709 & 0.875 & 0.105 & 25.164 & 0.832 & 0.117 & \textbf{30.272} & \textbf{0.931} & \textbf{0.060} \\
Barn     & 19.264 & 0.567 & 0.382 & 19.327 & 0.560 & 0.354 & 21.623 & 0.625 & 0.280 & 19.003 & 0.534 & 0.458 & \textbf{22.863} & \textbf{0.657} & \textbf{0.262} \\
Church   & 26.825 & 0.869 & 0.142 & \textbf{31.046} & \textbf{0.942} & \textbf{0.073} & 25.960 & 0.831 & 0.206 & 27.308 & 0.885 & 0.133 & 29.246 & 0.922 & 0.092 \\
Family   & 25.033 & 0.845 & 0.164 & 27.801 & 0.889 & 0.124 & 28.663 & 0.899 & 0.119 & 27.147 & 0.896 & 0.119 & \textbf{29.079} & \textbf{0.918} & \textbf{0.104} \\
Francis  & 23.898 & 0.737 & 0.270 & 24.313 & 0.717 & \textbf{0.257} & 24.365 & 0.738 & 0.272 & 24.774 & 0.727 & 0.258 & \textbf{25.359} & \textbf{0.742} & 0.266 \\
Horse    & 17.556 & 0.598 & 0.281 & 17.806 & 0.606 & 0.279 & \textbf{28.364} & \textbf{0.906} & \textbf{0.102} & 17.864 & 0.609 & 0.280 & 26.190 & 0.854 & 0.140 \\
Ignatius & 21.762 & 0.691 & 0.213 & 22.465 & \textbf{0.770} & \textbf{0.161} & \textbf{24.315} & 0.714 & 0.262 & 21.976 & 0.704 & 0.195 & 23.135 & 0.768 & 0.172 \\
Museum   & 16.194 & 0.523 & 0.437 & 17.480 & 0.540 & 0.420 & 23.136 & 0.737 & 0.214 & 18.397 & 0.503 & 0.393 & \textbf{23.821} & \textbf{0.773} & \textbf{0.201} \\
\midrule
Mean     & 21.931 & 0.707 & 0.252 & 23.232 & 0.734 & 0.222 & 25.267 & 0.791 & 0.195 & 22.704 & 0.711 & 0.244 & \textbf{26.246} & \textbf{0.821} & \textbf{0.162} \\
\bottomrule
\end{tabular}
\end{table*}

\begin{figure*}[t]
    \centering
    \includegraphics[width=\textwidth]{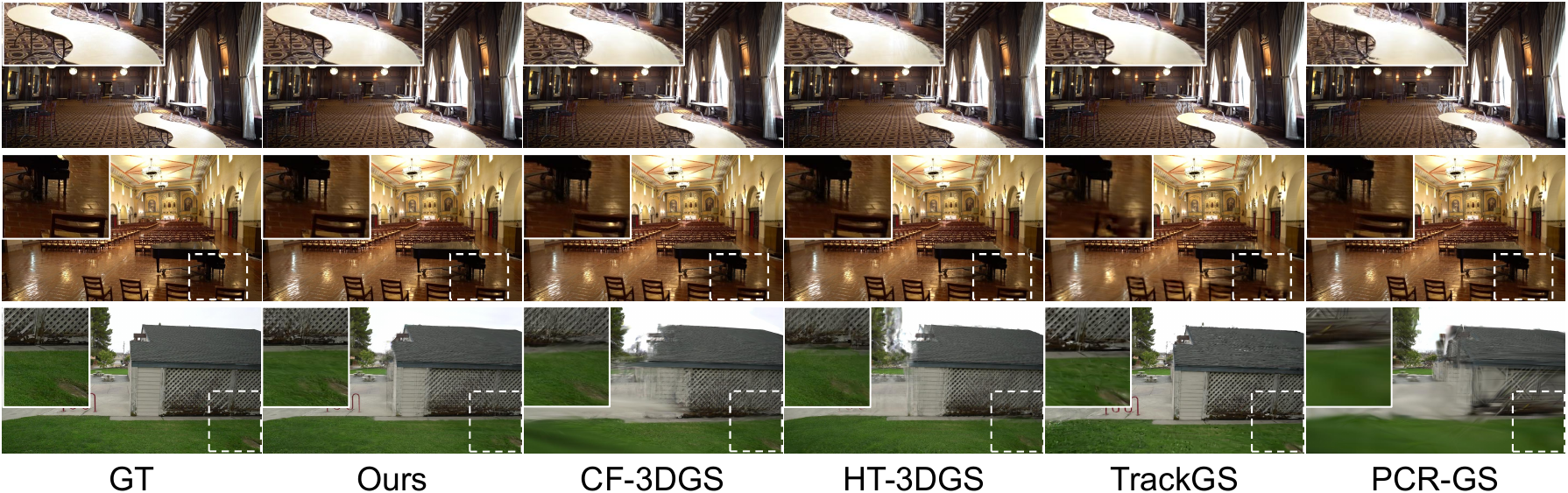}
    \caption{
        \textbf{Qualitative comparison of novel-view synthesis
        on Tanks and Temples.} CrossFit-GS preserves fine textures and sharp structural boundaries while reducing rendering artifacts.
    }
    \label{fig:tanks_visualization}
\end{figure*}

\begin{table*}[t]
\centering
\small
\caption{Quantitative NVS results on selected CO3D-V2 sequences. Evaluated using PSNR~$\uparrow$, SSIM~$\uparrow$, and LPIPS~$\downarrow$. Best results are bolded.}
\label{tab:co3d_nvs}
\setlength{\tabcolsep}{2.7pt}
\begin{tabular}{l|ccc|ccc|ccc|ccc|ccc}
\toprule
\multirow{2}{*}{Sequence} & \multicolumn{3}{c|}{CF-3DGS~\cite{cf-3dgs}} & \multicolumn{3}{c|}{HT-3DGS~\cite{ji2025sfm}} & \multicolumn{3}{c|}{PCR-GS~\cite{wei2025pcr}} & \multicolumn{3}{c|}{TrackGS~\cite{shi2026trackgs}} & \multicolumn{3}{c}{\textbf{Ours}} \\
& PSNR & SSIM & LPIPS & PSNR & SSIM & LPIPS & PSNR & SSIM & LPIPS & PSNR & SSIM & LPIPS & PSNR & SSIM & LPIPS \\
\midrule
apple\_23361    & 31.758 & 0.899 & 0.134 & 30.477 & 0.859 & 0.163          & 34.305 & 0.935 & 0.106 & 34.059 & 0.935 & 0.103          & \textbf{35.034} & \textbf{0.939} & \textbf{0.101} \\
hydrant\_108483 & 21.852 & 0.724 & 0.314 & 22.337 & 0.759 & 0.279          & 22.129 & 0.722 & 0.317 & 20.822 & 0.658 & 0.320          & \textbf{23.432} & \textbf{0.779} & \textbf{0.263} \\
plant\_50907    & 25.931 & 0.744 & 0.310 & 27.463 & 0.795 & 0.283          & 25.554 & 0.733 & 0.316 & 23.178 & 0.643 & 0.389          & \textbf{30.543} & \textbf{0.859} & \textbf{0.235} \\
teddybear\_38541& 32.048 & 0.862 & 0.334 & 33.957 & 0.900 & \textbf{0.279} & 32.059 & 0.862 & 0.333 & 33.509 & 0.865 & 0.288          & \textbf{34.732} & \textbf{0.903} & 0.290 \\
bench\_110109   & 32.509 & 0.894 & 0.144 & 30.302 & 0.844 & 0.195          & 31.397 & 0.856 & 0.164 & 27.415 & 0.713 & 0.277          & \textbf{34.595} & \textbf{0.933} & \textbf{0.109} \\
suitcase\_8645  & 28.587 & 0.856 & 0.269 & 32.020 & 0.920 & 0.224          & 30.862 & 0.891 & 0.253 & \textbf{32.321} & \textbf{0.923} & \textbf{0.143} & 30.019 & 0.900 & 0.244 \\
\midrule
Mean            & 28.781 & 0.830 & 0.251 & 29.426 & 0.846 & 0.237          & 29.384 & 0.833 & 0.248 & 28.551 & 0.789 & 0.253          & \textbf{31.393} & \textbf{0.886} & \textbf{0.207} \\
\bottomrule
\end{tabular}
\end{table*}

\subsection{Frozen-Map Cross-Fitted Pose Refinement}
\label{sec:crosspose}

Having obtained the local proposal $\widetilde{T}_i$, CrossFit-GS refines the target camera pose against the opposite submap $\mathcal{G}^{(b(i))}$. Throughout refinement, all Gaussian parameters in $\mathcal{G}^{(b(i))}$ are frozen, and the pose is initialized as $T_i^{(0)}=\widetilde{T}_i$. Consequently, the opposite submap cannot deform to reduce the target-frame loss; only the camera transformation is updated. To ensure robust convergence and reliable trajectory updates, we introduce spatial, photometric, and validation controls built upon this cross-fit objective.

\subsubsection{Initial Visibility Support and Coverage Gating}
We first evaluate the spatial overlap between the candidate pose and the opposite map using the rendered accumulated opacity:
\begin{align}
    V_i(p) &= \mathbf{1}\!\left[A(\mathcal{G}^{(b(i))}, \widetilde{T}_i; K)(p) > \tau_\alpha\right], \\
    c_i^{(0)} &= |\Omega_i|^{-1}\sum_{p\in\Omega_i}V_i(p),
    \label{eq:visibility}
\end{align}
where $\Omega_i$ denotes the image pixel grid and $\tau_\alpha$ is the opacity threshold. The ratio $c_i^{(0)}$ measures initial view coverage. To prevent unconstrained pose optimization on unmapped scene regions, we apply an initial coverage gate: If $c_i^{(0)}<\tau_{\mathrm{cov}}$, cross-map optimization is bypassed, we set $\widehat{T}_i=\widetilde{T}_i$, and proceed directly to the owner submap update. Otherwise, $V_i$ is fixed as the valid evaluation domain for frame $i$.

\subsubsection{Checkerboard Masking and Pose Optimization}

Within the valid visibility support $V_i$, we introduce checkerboard masking (CM) using a deterministic binary mask $H_i(p)$:
\begin{equation}
    H_i(p) = \begin{cases}
    1, & \text{if } (x_p+y_p)\bmod 2=0, \\
    0, & \text{otherwise}.
    \end{cases}
    \label{eq:checkerboard}
\end{equation}
This mask divides the visible pixels into two spatially interleaved and disjoint sets: optimization pixels $V_iH_i$, which provide gradients for pose refinement, and validation pixels $V_i(1-H_i)$, which are excluded from gradient computation.

For pose refinement, the optimizable camera pose $T$ is initialized from $\widetilde{T}_i$ and optimized over the support region $V_i H_i$:
\begin{equation}
\begin{split}
    \mathcal{L}_{\mathrm{pose}}(T)
    ={}&
    \frac{1}{N_i}
    \sum_{p\in\Omega_i}
    V_i(p)H_i(p)\,
    \ell_{\mathrm{smooth}}
    \!\left(\mathcal{R}_{p}(T),I_{i,p}\right) \\
    &+
    \lambda_{\mathrm{anch}}D(T,\widetilde{T}_i),
\end{split}
\label{eq:poseobjective}
\end{equation}
where $N_i=\max\!\left(1,\sum_{p\in\Omega_i}V_i(p)H_i(p)\right)$ is the number of valid optimization pixels, $\mathcal{R}(T)=\mathcal{R}(\mathcal{G}^{(b(i))},T;K)$, and $\mathcal{L}_{\mathrm{smooth}}$ denotes the channel-averaged Smooth-$L_1$ loss. The spatial anchor regularizer $D(T, \widetilde{T}_i)$ penalizes rotational and scale-normalized translational deviations from proposal $\widetilde{T}_i$:
\begin{equation}
    D(T, \widetilde{T}_i) = \frac{1}{9}\bigl\| R(T) - R(\widetilde{T}_i) \bigr\|_F^2 + \frac{1}{3 r_i^2}\bigl\| t(T) - t(\widetilde{T}_i) \bigr\|_2^2,
    \label{eq:anchor_reg}
\end{equation}
with $r_i$ representing the spatial radius of submap $\mathcal{G}^{(b(i))}$ to maintain scale consistency across diverse scenes.

Conversely, the complementary region $V_i(1-H_i)$ serves purely as a held-out validation set to evaluate pose candidates:
\begin{equation}
    \mathcal{L}_{\mathrm{val}}(T)
    =
    \frac{1}{N_i^{\mathrm{val}}}
    \sum_{p\in\Omega_i}
    V_i(p)\bigl(1-H_i(p)\bigr)
    \ell_{\mathrm{smooth}}
    \!\left(\mathcal{R}_p(T),I_{i,p}\right),
    \label{eq:validation}
\end{equation}
where $N_i^{\mathrm{val}}
=\max\!\left(1,\sum_{p\in\Omega_i}V_i(p)(1-H_i(p))\right)$. Without back-propagating gradients, $\mathcal{L}_{\mathrm{val}}(T)$ provides unbiased photometric evidence to guide candidate selection and acceptance.

\subsubsection{Retention Guard and Accept-or-Keep Decision}
During pose optimization, we continuously track the visible support fraction $q_i(T)$ to prevent pose divergence into unobserved space:
\begin{equation}
    q_i(T) = \frac{\sum_p V_i(p) V_i^T(p)}{\max\!\left(1, \sum_p V_i(p)\right)},
    \label{eq:retention}
\end{equation}
where $V_i^T$ denotes the synthesized opacity mask under current pose $T$. Optimization terminates early if $q_i(T)$ falls below threshold $\tau_{\mathrm{ret}}$. Among generated pose states, we select $T_i^{\mathrm{best}}$ that minimizes error on the validation set $\mathcal{L}_{\mathrm{val}}(T)$. The candidate pose is committed to the trajectory if and only if it yields a sufficient error reduction over proposal $\widetilde{T}_i$:
\begin{equation}
    \widehat{T}_i = \begin{cases}
    T_i^{\mathrm{best}}, & \text{if } \mathcal{L}_{\mathrm{val}}(T_i^{\mathrm{best}}) + \epsilon_{\mathrm{gain}} < \mathcal{L}_{\mathrm{val}}(\widetilde{T}_i), \\
    \widetilde{T}_i, & \text{otherwise}.
    \end{cases}
    \label{eq:accept}
\end{equation}

Upon committing $\widehat{T}_i$, frame $I_i$ updates its owner submap $\mathcal{G}^{(a(i))}$ while holding extrinsics fixed. Primitive parameters are optimized via the standard combined photometric objective:
\begin{equation}
    \mathcal{L}_{\mathrm{map}} = (1 - \lambda_{\mathrm{dssim}}) \mathcal{L}_1(\mathcal{R}, I_i) + \lambda_{\mathrm{dssim}} \mathcal{L}_{\mathrm{D-SSIM}}(\mathcal{R}, I_i),
    \label{eq:map_loss}
\end{equation}
thereby maintaining strict observation isolation across both submap partitions.

\subsection{Deferred Global Scene Reconstruction}
\label{sec:fusion}

Upon completing the streamwise trajectory optimization, the intermediate submaps $\mathcal{G}^{(0)}$ and $\mathcal{G}^{(1)}$ have fulfilled their purpose and are discarded. Under our deferred-commitment protocol, all verified camera extrinsics $\{\widehat{T}_i\}_{i=0}^{N-1}$ are permanently frozen, converting the optimized trajectory into immutable geometric anchors. A single global Gaussian representation $\mathcal{G}_{\mathrm{final}}$ is then initialized from $I_0$ and trained from scratch across all available views $\{I_i\}_{i=0}^{N-1}$ by minimizing the global photometric error:
\begin{equation}
    \mathcal{L}_{\mathrm{final}} = \frac{1}{N} \sum_{i=0}^{N-1} \mathcal{L}_{\mathrm{map}}\!\left(\mathcal{R}(\mathcal{G}_{\mathrm{final}}, \widehat{T}_i; K), I_i\right).
    \label{eq:final_loss}
\end{equation}
By deferring full-capacity scene modeling until the trajectory is locked, the final stage prevents early pose errors from being permanently encoded in the output representation and enables high-quality novel-view synthesis.

\begin{figure*}[t]
    \centering
    \includegraphics[width=\textwidth]{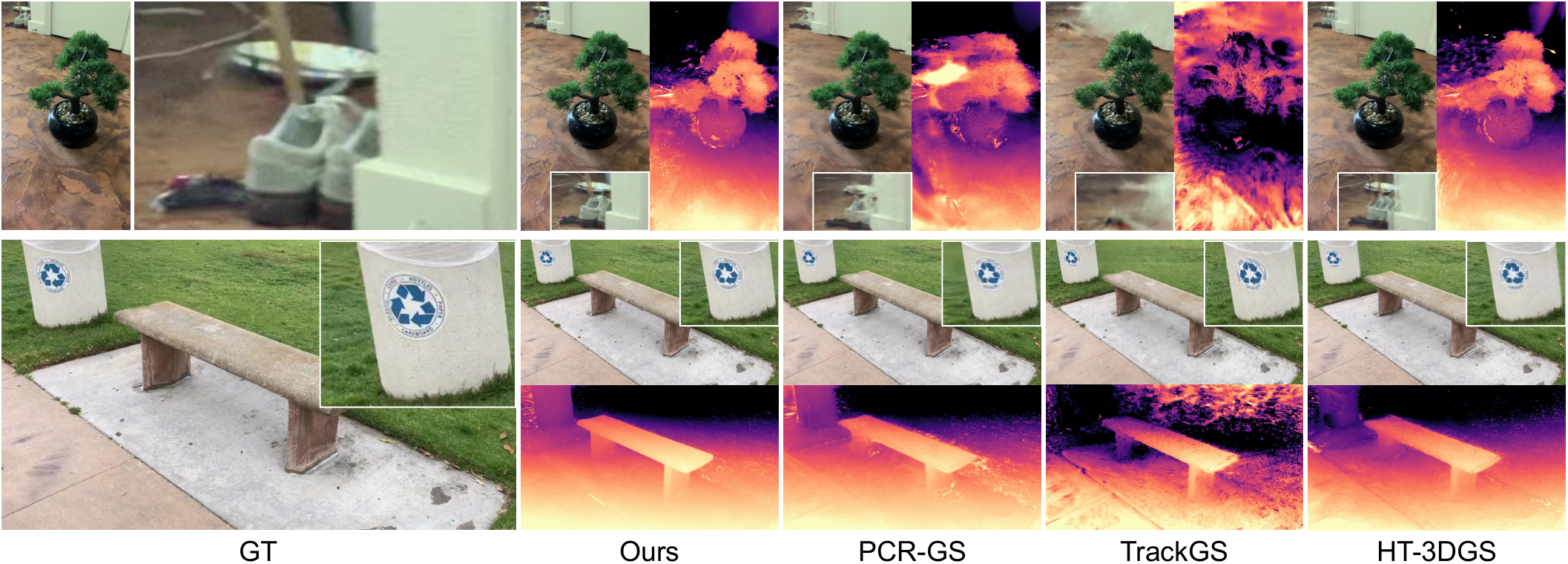}
    \caption{
        \textbf{Qualitative comparison of novel-view synthesis
        on CO3D-V2.} CrossFit-GS delivers highly accurate geometry and sharp visual details under rapid camera rotations, effectively suppressing blur and floating artifacts.
    }
    \label{fig:co3d}
\end{figure*}

\begin{figure*}[t]
    \centering
    \includegraphics[
        width=0.9\textwidth
    ]{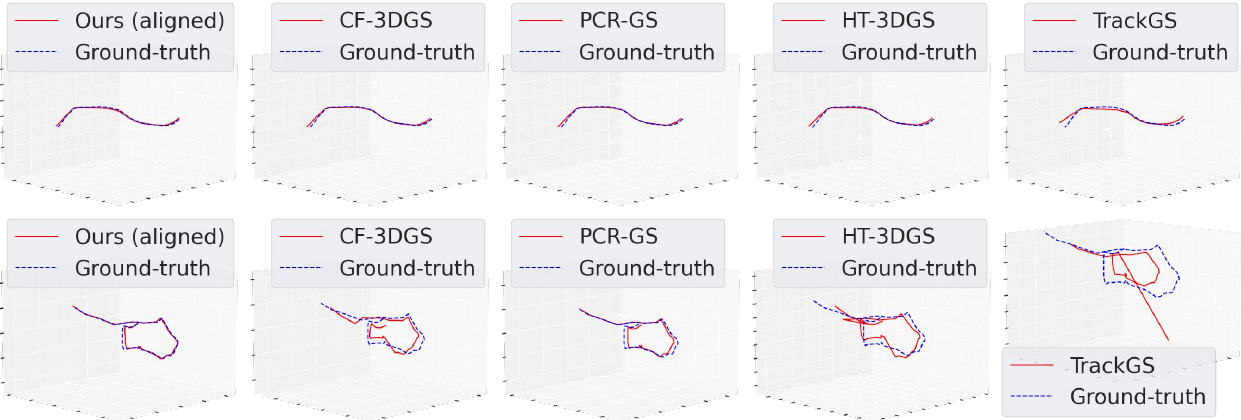}
    \caption{
        \textbf{Visualization of estimated camera trajectories.}
        The top row shows the Ignatius sequence from Tanks and Temples,
        and the bottom row shows the apple\_23361 sequence from CO3D-V2. Red solid lines denote estimated trajectories, while blue dashed lines indicate the ground truth.
    }
    \label{fig:trajectory}
\end{figure*}
\begin{table}[t]
\centering
\small
\caption{Quantitative camera pose estimation results on Tanks and Temples (stride 5) and selected CO3D-V2 sequences. Evaluated using $\text{RPE}_t \downarrow (\times 10^{-2})$, $\text{RPE}_r \downarrow (^\circ)$, and $\text{ATE} \downarrow$. Best results are bolded.}
\label{tab:pose_combined}
\setlength{\tabcolsep}{4.5pt}
\begin{tabular}{l|ccc|ccc}
\toprule
\multirow{2}{*}{Method} & \multicolumn{3}{c|}{Tanks and Temples} & \multicolumn{3}{c}{CO3D-V2} \\
& $\text{RPE}_t$ & $\text{RPE}_r$ & $\text{ATE}$ & $\text{RPE}_t$ & $\text{RPE}_r$ & $\text{ATE}$ \\
\midrule
CF-3DGS~\cite{cf-3dgs}     & 1.327 & 0.695 & 0.024 & 0.515 & 0.336 & 0.018 \\
HT-3DGS~\cite{ji2025sfm}   & 1.229 & 0.667 & 0.022 & 0.460 & 0.324 & 0.019 \\
TrackGS~\cite{shi2026trackgs} & 0.600 & 0.348 & 0.014 & 0.992 & 0.782 & 0.041 \\
PCR-GS~\cite{wei2025pcr}   & 1.440 & 0.596 & 0.041 & 0.453 & 0.331 & 0.014 \\
\midrule
\textbf{Ours}              & \textbf{0.571} & \textbf{0.347} & \textbf{0.010} & \textbf{0.344} & \textbf{0.323} & \textbf{0.009} \\
\bottomrule
\end{tabular}
\end{table}

\section{Experiments}
\label{sec:experiments}

We evaluate CrossFit-GS on challenging unposed benchmarks to assess both novel view synthesis fidelity and camera pose recovery accuracy.

\subsection{Experimental Setup}
\label{sec:experimental_setup}

\textbf{Datasets.}
Following previous benchmark evaluation protocols \cite{wei2025pcr, ji2025sfm, cf-3dgs, shi2026trackgs}, we evaluate on Tanks and Temples~\cite{tanks} using a frame-subsampling stride of 5 to simulate challenging wide-baseline camera motion. Additionally, we benchmark on CO3D-V2~\cite{co3d} across representative scenes featuring diverse geometric and textural complexities. In line with standard evaluation setups, every 8th frame is held out as test views for novel view synthesis.

\textbf{Metrics.}
Novel view synthesis is evaluated using PSNR, SSIM~\cite{ssim}, and LPIPS~\cite{lpips}. Camera trajectory estimation is aligned via $\mathrm{Sim}(3)$ Umeyama alignment and evaluated using Absolute Trajectory Error (ATE) as well as Relative Pose Errors for translation ($\text{RPE}_t$) and rotation ($\text{RPE}_r$)~\cite{barf, nope-nerf}.

\textbf{Implementation Details.}
Training sequences are split into alternating A/B folds ($i \bmod 2$). Local relative registration uses $1{,}000$ iterations for reference view fitting and $300$ for target tracking. Cross-map pose optimization uses Adam ($1.0 \times 10^{-3}$) for up to $100$ iterations, gated by $20\%$ initial alpha coverage and checkerboard-masked validation. A pose update is committed if validation loss improves by $> 1.0 \times 10^{-5}$ while retaining $\ge 50\%$ visible coverage. Frozen poses are then used to train a unified global Gaussian map for $500 \times N$ steps from scratch.

\subsection{Novel View Synthesis Evaluation}
\label{sec:nvs_eval}

Table~\ref{tab:tanks_nvs} and Table~\ref{tab:co3d_nvs} report the quantitative visual quality on the Tanks and Temples and CO3D-V2, respectively. CrossFit-GS consistently outperforms existing baselines across both datasets. On Tanks and Temples, our method achieves a mean PSNR of $26.246$~dB, representing a $4.315$~dB gain over CF-3DGS and a $0.979$~dB advantage over TrackGS. Similarly, on CO3D-V2, it yields an overall mean PSNR of $31.393$~dB, showing marked improvements on complex scenes such as plant\_50907.

These gains validate our observation-isolated design. Existing progressive methods often absorb residual pose errors into flexible Gaussian primitives, causing geometric artifacts and texture blur (Fig.~\ref{fig:tanks_visualization} and~\ref{fig:co3d}). By refining poses against frozen out-of-fold submaps and deferring global map updates until trajectories are fixed, CrossFit-GS prevents error propagation and renders sharper geometry and fine details.

\subsection{Trajectory Estimation Evaluation}
\label{sec:trajectory_eval}

Table~\ref{tab:pose_combined} reports the quantitative trajectory errors across both benchmarks. CrossFit-GS delivers the highest camera tracking precision, reducing the ATE on Tanks and Temples from $0.024$ for CF-3DGS to $0.010$, which represents a $58.3\%$ improvement and surpasses TrackGS ($0.014$). On CO3D-V2, our method achieves a mean ATE of $0.009$ alongside a mean $\mathrm{RPE}_t = 0.344$.

The trajectory visual comparisons in Fig.~\ref{fig:trajectory} demonstrate that CrossFit-GS preserves global motion paths under complex camera movements. This gain stems from the interaction between local pose initialization and out-of-fold refinement. The local stage provides short-range motion estimates, while cross-fit refinement checks those estimates against the multi-view context of the opposing submap. Supported by spatial anchoring and the accept-or-keep verification, these safeguards limit local registration errors and yield stable long-term trajectories.

\begin{table}[t]
\centering
\small
\caption{Ablation study on Tanks and Temples. Evaluated across pose metrics ($\text{RPE}_t \downarrow \times 10^{-2}$, $\text{RPE}_r \downarrow ^\circ$, $\text{ATE} \downarrow$) and NVS metrics ($\text{PSNR} \uparrow$, $\text{SSIM} \uparrow$, $\text{LPIPS} \downarrow$). Best results are bolded.}
\label{tab:ablation}
\resizebox{\linewidth}{!}{%
\setlength{\tabcolsep}{3pt}
\begin{tabular}{l|ccc|ccc}
\toprule
\multirow{2}{*}{Variant} & \multicolumn{3}{c|}{Pose} & \multicolumn{3}{c}{NVS} \\
& $\text{RPE}_t$ & $\text{RPE}_r$ & $\text{ATE}$ & $\text{PSNR}$ & $\text{SSIM}$ & $\text{LPIPS}$ \\
\midrule
Local tracking only     & 1.327 & 0.695 & 0.024 & 21.931 & 0.707 & 0.252 \\
Naive cross-fit         & 0.636 & 0.381 & 0.017 & 25.486 & 0.801 & 0.174 \\
w/o Coverage Gate       & 0.643 & 0.370 & 0.016 & 25.835 & 0.813 & 0.170 \\
w/o Retention Guard     & 0.578 & 0.367 & 0.014 & 25.909 & 0.812 & 0.167 \\
w/o Checkerboard        & 0.612 & 0.362 & 0.015 & 25.733 & 0.810 & 0.167 \\
w/o Accept-or-Keep      & 0.652 & 0.370 & 0.016 & 25.807 & 0.812 & 0.169 \\
\midrule
\textbf{Full CrossFit-GS} & \textbf{0.571} & \textbf{0.347} & \textbf{0.010} & \textbf{26.246} & \textbf{0.821} & \textbf{0.162} \\
\bottomrule
\end{tabular}%
}
\end{table}

\begin{table}[t]
\centering
\small
\caption{Efficiency comparison on the first 50 frames of selected CO3D-V2 sequences. Total runtime is measured in minutes (m).}
\label{tab:runtime}
\setlength{\tabcolsep}{10pt}
\begin{tabular}{l|c}
\toprule
Method & Total Time (min) $\downarrow$ \\
\midrule
HT-3DGS~\cite{ji2025sfm}    & 91.0 \\
TrackGS~\cite{shi2026trackgs} & 88.0 \\
CF-3DGS~\cite{cf-3dgs}      & 64.0 \\
\midrule
\textbf{Ours}               & 67.5 \\
\bottomrule
\end{tabular}
\end{table}

\subsection{Ablation Studies}
\label{sec:ablation}

\subsubsection{Contribution of Each Component}
\label{sec:ablation_components}

To quantify the contribution of each core component in CrossFit-GS, we conduct systematic ablation experiments on Tanks and Temples (Table~\ref{tab:ablation}).

Replacing local tracking with a naive cross-map optimization scheme (Naive cross-fit) yields a substantial performance leap, reducing ATE from $0.024$ to $0.017$ and increasing PSNR from $21.931$~dB to $25.486$~dB. This validates that optimizing target camera poses against an out-of-fold frozen submap successfully breaks the target-to-map self-fitting loop.

Spatially interleaving optimization and validation pixels via checkerboard masking ensures unbiased candidate evaluation; removing it (w/o Checkerboard) degrades both ATE ($0.015$) and PSNR ($25.733$~dB). Similarly, disabling the accept-or-keep mechanism (w/o Accept-or-Keep) increases tracking error, confirming the necessity of filtering out unhelpful pose updates using held-out photometric evidence.

Boundary and coverage controls provide critical safeguards against divergence. Omitting the coverage gate (w/o Coverage Gate) impairs pose alignment ($\text{ATE}=0.016$) under limited spatial overlap, while removing the retention guard (w/o Retention Guard) drops PSNR to $25.909$~dB due to pose drift into unobserved space. Combining all components into Full CrossFit-GS achieves the best trajectory accuracy and NVS fidelity.

\begin{table}[t]
\centering
\small
\caption{Diagnostic ablation of the self-explanation pathway on Tanks and Temples. \textbf{Target-Used}: whether the target image is used to train the reference map before pose refinement. \textbf{Map Frozen}: whether the reference map parameters are frozen during pose refinement.}
\label{tab:self_explanation_ablation}
\resizebox{\linewidth}{!}{%
\setlength{\tabcolsep}{3pt}
\begin{tabular}{l|cc|c|c|c}
\toprule
Variant & Target-Used & Map Frozen & Validation Loss$\downarrow$ & ATE$\downarrow$ & PSNR$\uparrow$ \\
\midrule
TCTM                 & Yes & No  & 0.000849          & 0.028          & 23.702 \\
TCFM                 & Yes & Yes & \textbf{0.000850} & 0.027          & 23.789 \\
TETM                 & No  & No  & 0.005092          & 0.029          & 23.305 \\
\midrule
\textbf{TEFM (Ours)} & No  & Yes & 0.032814          & \textbf{0.010} & \textbf{26.246} \\
\bottomrule
\end{tabular}%
}
\end{table}

\subsubsection{Runtime Efficiency}
\label{sec:runtime}

We evaluate runtime efficiency on a single NVIDIA RTX 4090 GPU across the first 50 frames of CO3D-V2 sequences, with results summarized in Table~\ref{tab:runtime}. All measurements account for both mapping and pose optimization time.

Despite the additional cross-map refinement, CrossFit-GS remains close to
CF-3DGS in runtime because it avoids replay during progressive estimation
and trains the final global representation only once. It is also faster than
HT-3DGS and TrackGS, which require hierarchical map consolidation and
long-horizon joint optimization, respectively.

\subsubsection{Diagnosing Target-to-Map Self-Explanation}
\label{sec:diag_self_explanation}

To directly test our core hypothesis, we evaluate four variants across two orthogonal factors—whether the reference map incorporates the target view prior to refinement, and whether it remains frozen during optimization: Target-conditioned trainable (TCTM), Target-conditioned frozen (TCFM), Target-excluded trainable (TETM), and Target-excluded frozen (TEFM, ours).

Table~\ref{tab:self_explanation_ablation} reveals a clear mismatch between validation loss
and downstream accuracy. Target-conditioned maps obtain extremely low losses,
but freezing them barely changes ATE or PSNR because the target has already
been absorbed. Target exclusion alone is likewise insufficient when the map
can still adapt during pose optimization. Our target-excluded frozen variant
instead improves PSNR by 2.457~dB and reduces ATE by 63.0\% over the
target-conditioned frozen variant, despite having a substantially higher
validation loss. This inversion provides direct evidence of target-to-map
self-explanation.

\section{Conclusion}
\label{sec:conclusion}

We presented CrossFit-GS, a deferred-commitment framework designed for COLMAP-free 3DGS. By enforcing strict observation isolation and refining each candidate pose against a frozen out-of-fold submap, our approach effectively eliminates the target-to-map self-explanation loop and prevents pose errors from being absorbed into Gaussian primitives. Experiments on benchmarks demonstrate that CrossFit-GS achieves substantial improvements in both global trajectory accuracy and novel-view rendering. Furthermore, our diagnostic studies confirm that isolating target-frame observations during pose refinement is crucial for preventing trajectory drift and reconstructing geometrically faithful 3D representations.
{
    \small
    \bibliographystyle{ieeenat_fullname}
    \bibliography{main}
}

% WARNING: do not forget to delete the supplementary pages from your submission 
% \input{sec/X_suppl}

\end{document}